\documentclass[10pt,twocolumn,letterpaper]{article}

\usepackage{wacv}              
\usepackage[table]{xcolor}
\usepackage{array}
\usepackage{multirow}

\definecolor{wacvblue}{rgb}{0.21,0.49,0.74}
\usepackage[pagebackref,breaklinks,colorlinks,allcolors=wacvblue]{hyperref}

\title{NEURO-GUARD: Neuro-Symbolic Generalization and Unbiased Adaptive Routing for Diagnostics - Explainable Medical AI}

\author{
Midhat Urooj\\
Arizona State University\\
Tempe, AZ, USA\\
{\tt\small murooj@asu.edu}
\and
Ayan Banerjee\\
Arizona State University\\
Tempe, AZ, USA\\
{\tt\small abanerj3@asu.edu }
\and
Sandeep Gupta\\
Arizona State University\\
Tempe, AZ, USA\\
{\tt\small Sandeep.Gupta@asu.edu }
}

\begin{document}
\maketitle

\begin{abstract}
Accurate yet interpretable image-based diagnosis remains a central challenge in medical AI, particularly in settings with limited data, subtle visual patterns, and high-stakes clinical decisions. However, most current vision models produce black-box predictions with limited generalizability and poor real-world usability. We present \textbf{ NEURO-GUARD}, a novel framework that combines Vision Transformers (ViTs) with knowledge-guided reasoning to enhance \textbf{performance, transparency}, and \textbf{cross-domain generalization}.  NEURO-GUARD incorporates a \textbf{retrieval-augmented generation (RAG)} mechanism for language-driven self-verification, in which a large language model (LLM) iteratively generates, evaluates, and refines feature extraction code for medical images. By leveraging clinical guidelines and expert knowledge, this LLM-guided module progressively improves feature detection and classification, outperforming purely data-driven baselines. Extensive evaluations on diabetic retinopathy classification across four benchmark datasets (APTOS, EyePACS, Messidor-1, Messidor-2) show that  NEURO-GUARD improves accuracy by \textbf{6.2\%} over a ViT-only model (84.69\% vs. 78.4\% \cite{dosovitskiy2020}) and achieves a \textbf{5\%} gain in domain generalization. Further experiments on MRI-based seizure detection confirm its cross-domain robustness, consistently surpassing existing baselines. Notably,  NEURO-GUARD bridges the gap between symbolic medical reasoning and subsymbolic feature learning, demonstrating robust generalization across multiple datasets while achieving \textbf{state-of-the-art performance}.
\end{abstract}
\begin{figure}[t]  
    \centering
    \includegraphics[width=\columnwidth]{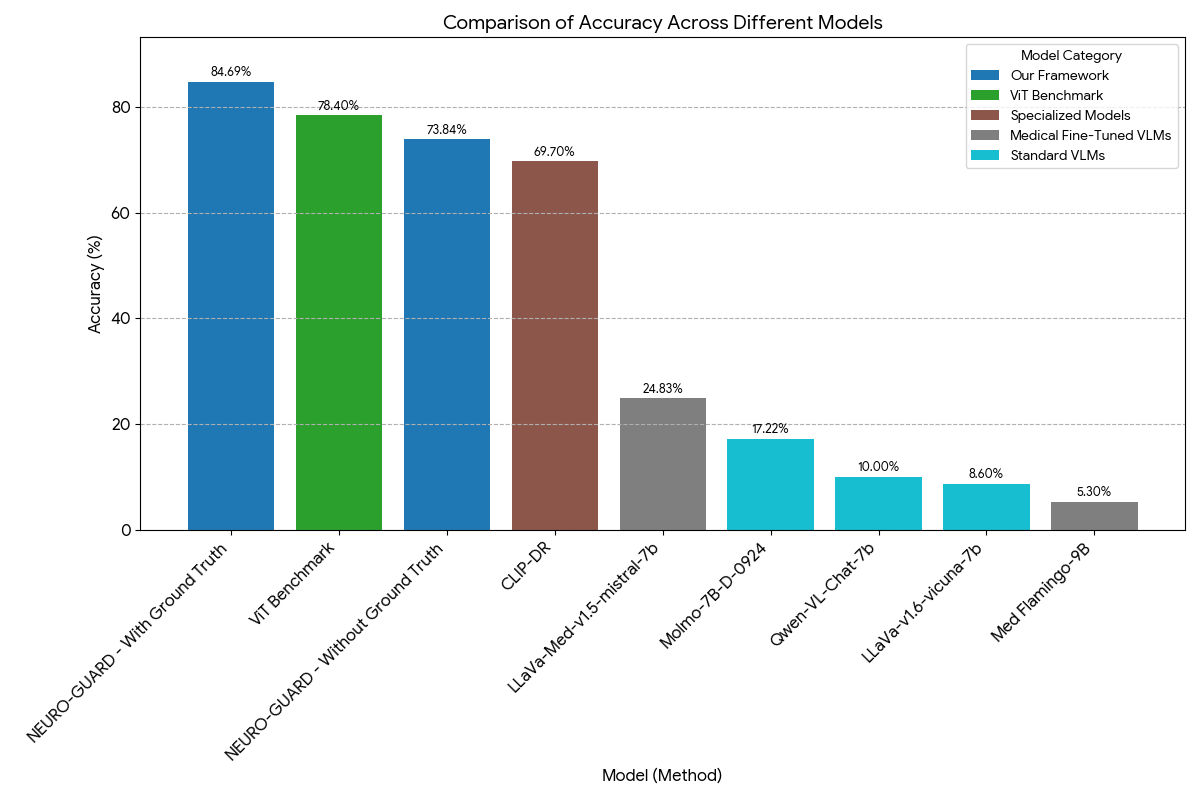} 
\caption{Performance comparison of existing models versus the NEURO-GUARD framework for 5-stage Diabetic Retinopathy classification.}

    \label{fig:deepxsoz_framework}
\end{figure}
 
\begin{figure*}[t]  
    \centering
    \includegraphics[width=\textwidth]{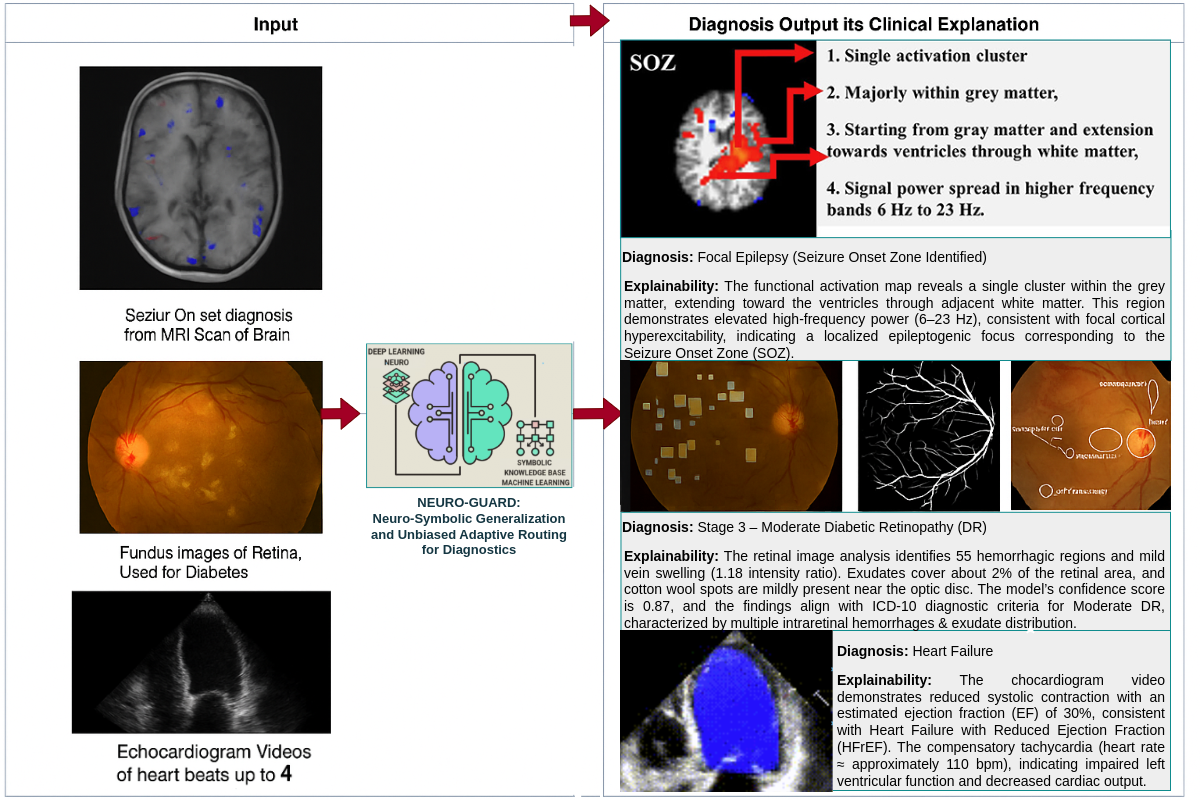} 
\caption{Overview of the NEURO-GUARD framework. The system integrates medical knowledge with multimodal imaging to enhance disease classification and provide clinically aligned, interpretable explanations with spatial localization. }

    \label{fig:deepxsoz_framework}
\end{figure*}

\section{Introduction}
Medical imaging plays a crucial role in disease diagnosis and treatment planning, particularly in conditions such as diabetic retinopathy (DR), tumor detection, and neurodegenerative disorders. Recent advances in deep learning, particularly Vision Transformers (ViTs) and Convolutional Neural Networks (CNNs), have significantly improved diagnostic accuracy \cite{dosovitskiy2020,simonyan2015}. However, their black-box nature limits clinical adoption due to a lack of interpretability, making it challenging for clinicians to validate AI-driven decisions. Additionally, these models suffer from domain shift vulnerabilities, struggling to generalize across imaging datasets with diverse acquisition protocols and patient demographics \cite{zhou2022,wu2022}. Given these challenges, an ideal medical AI framework should not only provide high accuracy but also generate clinically interpretable decisions by integrating structured domain knowledge into its reasoning process.

Existing explainability techniques, such as Gradient-weighted Class Activation Mapping (Grad-CAM) \cite{selvaraju2017} and Shapley Additive Explanations (SHAP) \cite{lundberg2017}, provide post-hoc feature attribution but remain static, heuristic-based, and disconnected from the model’s decision logic. Hybrid approaches incorporating attention mechanisms and uncertainty estimation attempt to improve interpretability \cite{wang2021,volpi2018}, but they fail to integrate structured medical knowledge, limiting their ability to generalize across datasets. Reinforcement learning (RL) and meta-learning frameworks \cite{mnih2015} enable adaptive learning, yet they lack mechanisms to ground AI decisions in clinical reasoning, reducing their reliability in real-world medical applications.

To address these limitations, we propose  NEURO-GUARD, a novel framework that fuses language-grounded reasoning with state-of-the-art visual recognition to enable intrinsically interpretable medical image diagnosis. In contrast to prior systems that only add interpretability after the fact,  NEURO-GUARD tightly integrates a clinical knowledge base and reasoning module into the model’s inference pipeline. This is achieved through a modular architecture combining a self-supervised ViT-based image encoder with a knowledge-guided language model that jointly analyzes images and textual information. Crucially, our approach leverages retrieval-augmented generation to dynamically draw on external biomedical sources (e.g., literature, guidelines) for case-specific knowledge, and uses an LLM-based code synthesis engine to translate this knowledge into executable image analysis steps. A prompt-driven self-verification loop, optimized via reinforcement learning, compels the model to iteratively check and refine its outputs greatly reducing hallucinations and aligning final predictions with clinical guidelines. Through this design,  NEURO-GUARD shifts interpretability from a post-hoc exercise to an intrinsic property of the model’s predictions, as the reasoning is conducted in natural language and grounded in real clinical criteria from the start. In essence, our framework bridges the gap between symbolic medical knowledge and subsymbolic vision features, allowing the model to explain why and how it arrives at a diagnosis in terms familiar to human experts as shown in Figure~\ref{fig:deepxsoz_framework}.

\begin{figure*}[ht]
    \centering
    \includegraphics[width=\textwidth]{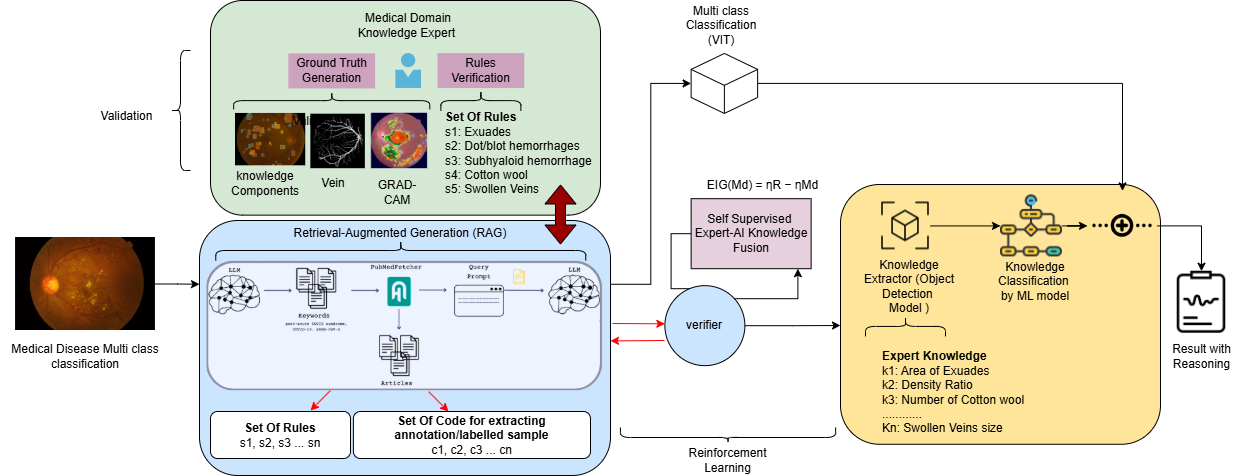}
    \caption{\textbf{ NEURO-GUARD Framework for Knowledge-Driven Medical AI.} The  NEURO-GUARD pipeline integrates RAG-based knowledge extraction, reinforcement learning-based self-verification, and multi-class classification.}
    \label{fig: NEURO-GUARD_framework}
\end{figure*}
\subsection{Contributions}

We design a \textbf{hybrid inference pipeline} that fuses deep-learning predictions with \textbf{knowledge-driven classifiers}, enabling transparency while maintaining high diagnostic performance. In contrast to existing VLMs and LLMs which often produce confident yet inaccurate and hallucinated explanations, as demonstrated in our Phase~1 and Phase~2 experiments. NEURO-GUARD grounds every decision in structured medical knowledge. We introduce the \textbf{first medical-imaging pipeline that transforms clinical guidelines and expert rules into executable, verifiable code}, integrating this symbolic reasoning directly with \textbf{Vision Transformer (ViT) feature learning}. Using a \textbf{multi-stage Retrieval-Augmented Generation (RAG) process} grounded in peer-reviewed medical literature and disease-specific protocols, our system constructs dynamic rule bases that guide pixel-level lesion detection with stronger clinical consistency. To further ensure reliability, we develop an \textbf{entropy-based reinforcement learning self-verification loop} that iteratively refines code-generated feature extractors, substantially reducing hallucinations and improving localization accuracy. Through comprehensive evaluations on \textbf{diabetic retinopathy datasets} (APTOS, EyePACS) as well as \textbf{MRI-based seizure detection}, we demonstrate that NEURO-GUARD achieves robust generalizability across diseases, imaging modalities, and clinical conditions, establishing a new direction for interpretable and trustworthy medical AI.

\section{Related Work}

\subsection{Interpretable Models vs. Black-Box Paradigms}
The trade-off between predictive performance and model transparency remains a defining challenge in medical artificial intelligence. While deep learning architectures, such as Convolutional Neural Networks (CNNs) \cite{simonyan2015} and Vision Transformers (ViTs) \cite{dosovitskiy2020}, have achieved state-of-the-art performance in diagnostic tasks, they operate as opaque ``black boxes.'' Early attempts to mitigate this opacity relied on post-hoc explanation methods like Grad-CAM \cite{selvaraju2017}, which highlight salient image regions. However, these methods have been criticized for providing fragile approximations that often fail to capture the model's true decision logic \cite{rudin2019}. Rudin \cite{rudin2019} argues that for high-stakes decisions, reliance on post-hoc explanations is insufficient and advocates for inherently interpretable architectures.

In response, Concept Bottleneck Models (CBMs) \cite{koh2020} were proposed to align latent representations with human-understandable concepts (e.g., ``bone spur'' or ``effusion'') prior to classification. While CBMs offer intrinsic interpretability, they typically require dense, costly concept annotations and often suffer from a performance gap compared to end-to-end models \cite{yan2023}. Alternative strategies employing reinforcement learning \cite{wu2022} offer dynamic adaptation but often lack mechanisms to ground learned representations in explicit domain knowledge. Unlike these static or purely data-driven approaches, our framework leverages the dynamic reasoning capabilities of Large Language Models (LLMs) to generate interpretable feature extractors without requiring pixel-level concept supervision.

\subsection{Vision-Language Models in Medicine}
The integration of linguistic knowledge into medical imaging has been accelerated by Vision-Language Models (VLMs). Building on the contrastive learning paradigm of CLIP \cite{radford2021}, domain-specific models such as MedCLIP \cite{wang2022} and Med-PaLM 2 \cite{singhal2023} have demonstrated remarkable capabilities in zero-shot classification and report generation. These models encode vast medical ontologies, enabling them to process complex diagnostic queries.

However, significant limitations persist. Recent studies indicate that generative medical agents frequently exhibit ``hallucinations,'' producing plausible but factually incorrect findings due to a lack of grounding in pixel-level evidence \cite{cohenwang2024}. Furthermore, most VLMs process information through separate, static encoders, losing the fine-grained feature alignment necessary for verifying subtle biomarkers \cite{zhang2022}. NEURO-GUARD addresses this by moving beyond static embeddings, using an agentic framework to actively query and verify visual features.

\subsection{Neuro-Symbolic and Agentic Frameworks}
To bridge the gap between symbolic reasoning and subsymbolic perception, recent works in general computer vision have proposed ``code-as-policy'' approaches. Systems such as VisProg \cite{gupta2023} and ViperGPT \cite{suris2023} utilize LLMs to decompose complex visual queries into executable Python programs, invoking vision primitives to solve tasks without specific training. These frameworks demonstrate that LLMs can act as reasoning engines to orchestrate vision modules effectively.

Despite their success in general domains, these agents lack the specialized clinical knowledge required for medical diagnostics. General-purpose code generation often fails to synthesize the precise, domain-specific subroutines needed to detect pathological features (e.g., distinguishing ``microaneurysms'' from ``hemorrhages''). Our work extends the neuro-symbolic paradigm by integrating Retrieval-Augmented Generation (RAG) with code generation, synthesizing expert clinical guidelines into executable logic that is both performant and verifiable.
\section{Proposed method}

Our proposed pipeline,  NEURO-GUARD, autonomously identifies and refines domain-specific features by integrating symbolic medical knowledge with subsymbolic learning, ensuring strong alignment with clinical standards. As depicted in Figure~\ref{fig: NEURO-GUARD_framework},  NEURO-GUARD employs a Retrieval-Augmented Generation (RAG) mechanism to extract structured medical knowledge from sources such as PubMed and clinical guidelines. For instance, in the context of diabetic retinopathy (DR), the system retrieves rules indicating that hemorrhages, exudates, and swollen veins are key visual markers. This retrieved knowledge is passed to a large language model (LLM) in a multi-prompt sequence. In Prompt 1, the LLM consolidates disease-specific information into a structured clinical rule base, detailing relevant features and diagnostic criteria. In Prompt 2, the LLM utilizes this rule base to generate executable Python code for feature detection, embedding reinforcement learning (RL) parameters to guide initial predictions. In Prompt 3, performance feedback including metrics such as Intersection-over-Union (IoU), precision, and recall is used to iteratively refine the generated code via RL, enhancing alignment with human annotations or model confidence. A self-verification module then evaluates whether the extracted features conform to expected clinical patterns using entropic reward signals.  Ultimately,  NEURO-GUARD produces interpretable and generalizable outputs by harmonizing LLM-driven symbolic reasoning with deep learning-based feature recognition.
\begin{figure}[t]  
    \centering
    \includegraphics[width=\columnwidth]{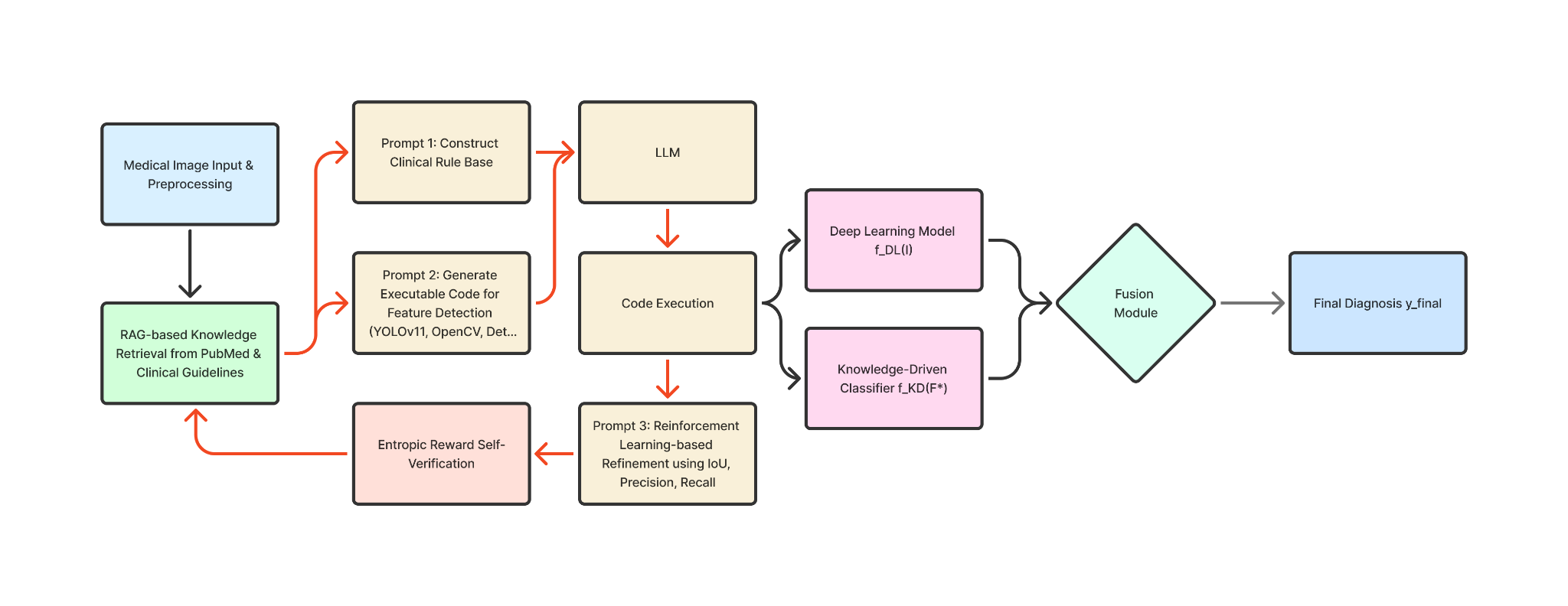} 
\caption{Flow diagram of the NEURO-GUARD framework }

    \label{fig:deepxsoz_framework}
\end{figure}

\subsection{Mathematical Formulation}

Given an input retinal image \(I\), we first apply a feature detection module \(\mathcal{F}\) to extract a comprehensive set of clinically relevant features \(F = \{f_1, f_2, \dots, f_n\}\). Each feature \(f_i\) corresponds to distinct anatomical or pathological markers, such as exudates, hemorrhages, and cotton wool spots. The feature set \(F\) is mapped into a vectorized representation \(v \in \mathbb{R}^d\) through a domain-specific embedding function \(\varphi: F \rightarrow \mathbb{R}^d\).

Simultaneously, we retrieve and structure clinical knowledge rules \(R\), which include visual representation rules \(R_V = \{r_{v1}, r_{v2}, \dots, r_{vm}\}\) and demographic knowledge rules \(R_D = \{r_{d1}, r_{d2}, \dots, r_{dl}\}\). These rules are dynamically extracted from authoritative clinical sources such as PubMed using a Retrieval-Augmented Generation (RAG) framework. Each visual rule \(r_{vi}\) is directly associated with a corresponding subset of image-derived features \(F_{rvi}\), and demographic rules \(r_{dj}\) are linked to patient-specific metadata features \(F_{rdj}\), forming an integrated knowledge extraction set:
\[
F = \{F_{rv1}, F_{rv2}, \dots, F_{rvm}, F_{rd1}, F_{rd2}, \dots, F_{rdl}\}
\]

\subsection{Visual Feature Extraction and Self-Verification}

Large Language Models (LLMs) are prompted to generate executable visual feature extraction codes \(C = \{ c_1, c_2, \dots, c_m \}\), each specifically tailored to detect features corresponding to visual rules \(r_{vi}\). These codes are instantiated through vision frameworks such as OpenCV, YOLOv11, or Detectron2. Extracted visual features \(X_i\) from an image \(I\) are quantitatively validated against ground truth annotations \(Y_i\) using Intersection-over-Union (IoU) metrics:
\[
\text{IoU}(X_i, Y_i) = \frac{|X_i \cap Y_i|}{|X_i \cup Y_i|}
\]

To enhance the reliability and accuracy of feature extraction, we introduce an entropic reward-based self-verification mechanism. The entropic gain \(E_i\) quantifies the reduction in uncertainty associated with feature extraction:
\[
E_i = -\sum_{k=1}^{K} p_k \log p_k
\]
where \(p_k\) denotes the empirical probability of correct extraction across \(K\) randomly selected validation images. If the entropy \(E_i\) is below a predefined threshold \(\tau\), iterative refinement of the extraction code \(c_i\) is performed using reinforcement learning (RL)-guided prompt tuning:
\[
c_i^{(t+1)} = \text{LLM}\left(E_i, c_i^{(t)}\right), \quad \text{until} \quad \max_{t} E_i^{(t)} \geq \tau
\]

Upon convergence, verified visual extraction codes \(C^*\) yield the optimized visual knowledge set \(F_{RV}^*\). Together with the demographic knowledge set \(F_{RD}\), they form the refined feature set:
\[
F^* = \{F_{RV}^*, F_{RD}\}
\]

\subsection{Classification and Decision Fusion}

The refined feature set \(F^*\) is input into a knowledge-driven classification model \(f_{\text{KD}}\), producing a predicted disease stage \(y_{\text{KD}} = f_{\text{KD}}(F^*)\). Concurrently, a deep learning-based model \(f_{\text{DL}}\), trained on extensive labeled retinal images, processes the input image \(I\), generating a prediction \(y_{\text{DL}} = f_{\text{DL}}(I)\) with associated confidence \(s_{\text{DL}}\).

The final diagnosis \(y_{\text{final}}\) is obtained by integrating both predictions through a fusion function \(f_{\text{fusion}}\):
\[
y_{\text{final}} =
\begin{cases}
y_{\text{DL}}, & \text{if } s_{\text{DL}} \geq s_{\text{KD}} \\
y_{\text{KD}}, & \text{otherwise.}
\end{cases}
\]

\subsection{Reinforcement Learning-Based Optimization}

To systematically optimize detection accuracy, we utilize an RL-based parameter tuning approach under two distinct feedback environments:

\paragraph{Supervised Artifact Detection - With Ground Truth Annotations} In this scenario, RL directly optimizes YOLO confidence thresholds:
\begin{itemize}
    \item \textbf{State Representation}: Discretized YOLO confidence threshold.
    \item \textbf{Action Space}: Adjustments by increments \{-0.05, 0, +0.05\}.
    \item \textbf{Reward Function}: IoU between predicted and annotated bounding boxes.
    \item \textbf{Q-learning Update}: Standard Q-learning algorithm with parameters \(\alpha = 0.1\), \(\gamma = 0.9\), determined via empirical sensitivity analysis.
\end{itemize}

\paragraph{Unsupervised Knowledge Extraction - Without Ground Truth Annotations} OpenCV parameters optimization is performed heuristically:
\begin{itemize}
    \item \textbf{State Representation}: Tuple of CLAHE clip limit and optic disc masking threshold.
    \item \textbf{Action Space}: Incremental parameter adjustments.
    \item \textbf{Reward Function}: Heuristic measure based on the accuracy of detected feature counts relative to clinically expected targets.
    \item \textbf{Training and Q-learning Update}: Mirrors the ground truth approach, maintaining consistent hyperparameter values for stability and generalization.
\end{itemize}

\begin{table*}[t]
    \centering
    \caption{
        Performance comparison of classification models on the Aptos and Eye Pacs datasets. 
        Models are tested with supervised knowledge and without supervised knowledge to assess their interpretability impact.
        The best-performing values per row are \textbf{bolded}, and the highest-performing model per dataset is \cellcolor{yellow} highlighted.
    }
    \label{tab:classification_results}
    \resizebox{\textwidth}{!}{%
    \begin{tabular}{llcccccccc}
        \toprule
        \textbf{Framework} & \textbf{Model} & \textbf{Dataset} & \textbf{Val. Acc.} & \textbf{Test Acc.} & \textbf{Precision} & \textbf{Recall} & \textbf{F1-Score} & \textbf{Weighted Prec.} & \textbf{Weighted F1} \\
        \midrule
        \multirow{5}{*}{\textbf{NEURO-GUARD (Unsupervised Knowledge Extraction)}} 
            & Logistic Regression & Aptos & 0.6019 & 0.6424 & 0.25 & 0.33 & 0.28 & 0.55 & 0.58 \\
            & \cellcolor{yellow} Random Forest & Aptos & 0.7038 & \textbf{0.7384} & 0.55 & 0.47 & 0.49 & 0.71 & 0.71 \\
            & SVM & Aptos & 0.6083 & 0.6556 & 0.26 & 0.34 & 0.28 & 0.56 & 0.59 \\
            & Gradient Boosting & Aptos & 0.7389 & 0.7252 & 0.51 & 0.44 & 0.44 & 0.70 & 0.70 \\
            & K-Nearest Neighbors & Aptos & 0.6369 & 0.6987 & 0.43 & 0.44 & 0.42 & 0.65 & 0.67 \\
        \midrule
        \multirow{5}{*}{\textbf{NEURO-GUARD (Unsupervised Knowledge Extraction)}} 
            & Logistic Regression & Eye Pacs & 0.6932 & 0.6991 & 0.45 & 0.40 & 0.42 & 0.68 & 0.69 \\
            & Random Forest & Eye Pacs & 0.7198 & 0.7342 & 0.50 & 0.44 & 0.47 & 0.73 & 0.74 \\
            & SVM & Eye Pacs & 0.7011 & 0.7103 & 0.48 & 0.42 & 0.45 & 0.72 & 0.72 \\
            & \cellcolor{yellow} Gradient Boosting & Eye Pacs & 0.7429 & \textbf{0.7465} & 0.55 & 0.48 & 0.52 & 0.75 & 0.75 \\
            & K-Nearest Neighbors & Eye Pacs & 0.7124 & 0.7206 & 0.50 & 0.43 & 0.46 & 0.72 & 0.73 \\
        \midrule
        \multirow{5}{*}{\textbf{NEURO-GUARD (Supervised Artifact Detection)}} 
            & Logistic Regression & Aptos & 0.7322 & 0.7732 & 0.59 & 0.49 & 0.49 & 0.77 & 0.75 \\
            & Random Forest & Aptos & 0.8005 & 0.7978 & 0.65 & 0.56 & 0.57 & 0.80 & 0.78 \\
            & SVM & Aptos & 0.7432 & 0.7814 & 0.59 & 0.50 & 0.50 & 0.77 & 0.75 \\
            & \cellcolor{yellow} Gradient Boosting & Aptos & 0.8415 & \textbf{0.8469} & 0.69 & 0.58 & 0.58 & 0.83 & 0.80 \\
            & K-Nearest Neighbors & Aptos & 0.7896 & 0.7814 & 0.63 & 0.56 & 0.55 & 0.78 & 0.76 \\
        \midrule
        \multirow{5}{*}{\textbf{NEURO-GUARD (Supervised Artifact Detection)}} 
            & Logistic Regression & Eye Pacs & 0.7354 & 0.7401 & 0.52 & 0.46 & 0.48 & 0.75 & 0.75 \\
            & Random Forest & Eye Pacs & 0.7651 & 0.7713 & 0.58 & 0.51 & 0.54 & 0.78 & 0.78 \\
            & SVM & Eye Pacs & 0.7423 & 0.7512 & 0.55 & 0.48 & 0.51 & 0.76 & 0.76 \\
            & \cellcolor{yellow} Gradient Boosting & Eye Pacs & 0.7742 & \textbf{0.7796} & 0.60 & 0.53 & 0.56 & 0.80 & 0.80 \\
            & K-Nearest Neighbors & Eye Pacs & 0.7545 & 0.7608 & 0.56 & 0.49 & 0.52 & 0.77 & 0.77 \\
        \bottomrule
    \end{tabular}
    }
\end{table*}

\begin{table}[h]
    \centering
    \caption{
        Performance comparison of our framework with DeepXSOZ, and CNN for seizure onset zone (SOZ) identification. 
        The benchmark results are based on anatomical MRI-based manual SOZ identification and surgical outcomes. 
        Our method achieves \textbf{83.27\%}, demonstrating strong generalization across multi-center datasets.
    }
    \label{tab:soz_comparison}
    \begin{tabular}{l c}
        \toprule
        \textbf{Method} & \textbf{SOZ Identification AUC } \\
        \midrule
        DeepXSOZ & 81.6 \\
        CNN & 46.1 \\
        \textbf{Our Framework} & \textbf{83.27} \\
        \bottomrule
    \end{tabular}
\end{table}

\section{Experimental Setup and Evaluation}
\label{sec:dr_classification}

Diabetic Retinopathy (DR) remains a leading cause of visual impairment, necessitating diagnostic methods that are not only accurate but also interpretable. While deep learning has achieved success in classification \cite{mlynarski2023}, clinical adoption requires the precise localization of pathological features such as exudates and hemorrhages to justify decisions. Our objective was to achieve this reasoning and explainability. However, our initial experiments with state-of-the-art foundation models revealed critical reliability gaps, motivating the development of the \textbf{NEURO-GUARD} framework. The experimental setup shows the full journey of experimental phases from getting results based on hallucinations to highly accurate results with correct Reasoning from our NEURO-GUARD framework.

\paragraph{Phase 1: Failure of Zero-Shot Vision-Language Models.}
We first evaluated off-the-shelf Vision-Language Models (VLMs) on the APTOS and EyePACS datasets. Inspired by the CARES benchmark for trustworthiness \cite{xia2024}, we tested zero-shot and few-shot capabilities. As detailed in Table \ref{tab:vlm_observations}, these models struggled significantly with medical granularity:
\begin{itemize}
    \item \textbf{CLIP \& MedCLIP:} These models relied on holistic semantics and failed to capture subtle lesions, often misclassifying severe DR as mild due to a lack of object-level grounding \cite{antaki2024}.
    \item \textbf{Grounding-DINO:} Produced frequent false positives by incorrectly labeling non-pathological artifacts (e.g., optic disc reflections) as lesions \cite{tang2022}.
    \item \textbf{InstructBLIP:} Lacked the spatial resolution required to detect small markers like microaneurysms.
\end{itemize}

\begin{table}[t]
\centering
\caption{Summary of VLM Observations for DR Classification}
\label{tab:vlm_observations}
\resizebox{\columnwidth}{!}{%
\begin{tabular}{lp{0.58\columnwidth}}
\toprule
\textbf{Model} & \textbf{Key Observations} \\
\midrule
Qwen-VL-Chat & Generated strong textual reasoning yet achieved only 10.4\% accuracy, notably below its 33.84\% CARES benchmark. \\
\midrule
LLaVa-Med-v1.5-mistral-7b & Improved classification accuracy to 24.83\% by explicitly defining features, but still prone to hallucination. \\
\midrule
LLaVa-v1.6-vicuna-7b & Performed admirably in captioning but failed in fine-grained feature detection, limiting reliability for DR classification. \\
\midrule
Molmo-7B-D-0924 & Demonstrated robust reasoning yet frequently hallucinated features, resulting in misclassifications. \\
\midrule
CogVLM & Excelled at descriptive captioning capabilities but lacked structured classification functionality. \\
\bottomrule
\end{tabular}
}
\end{table}

These findings align with recent studies on medical hallucinations, where models generate convincing but factually incorrect diagnostic captions \cite{gu2024, holland2024}.

\paragraph{Phase 2: Limitations of Direct LLM-Based Feature Extraction.}
Given the localization failures of VLMs, we explored using Large Language Models (LLMs) like Gemini, Claude, and GPT-4 for structured feature extraction. While LLMs excelled at reasoning, they struggled with direct visual grounding. For instance, Gemini often misclassified bright artifacts as exudates, and MedGPT produced numerous false positives despite domain-specific tuning. GPT-4 achieved the highest accuracy (73\%) through iterative bounding-box refinement but remained inconsistent \cite{wang2024}.

\paragraph{Phase 3: The NEURO-GUARD Solution.}
The failure of black-box models to provide reliable, non-hallucinated explanations prompted us to move beyond direct prediction. We propose \textbf{NEURO-GUARD}, an agentic framework where the LLM does not diagnose directly but instead acts as a logic engine. It generates deterministic Python code (e.g., OpenCV functions) to extract features based on clinical expert rules, ensuring that the diagnosis is grounded in verifiable image data rather than stochastic model outputs.We compare the performance of traditional deep learning models, VLMs, LLMs, and the specialized fine-tuned CLIP-DR model on diabetic retinopathy with the results of our proposed framework, which clearly demonstrates superior outcomes \ref{tab:results}.
\subsection{Localization and Clinical Interpretability Assessment for Diabetic Retinopathy (DR) }

To evaluate lesion-level interpretability, we use a YOLOv11 detector fine-tuned on 300 expert-annotated retinal images containing exudates, hemorrhages, cotton-wool spots, and venous abnormalities. The clinical categories used for annotation are first retrieved through our RAG module from PubMed and then verified by ophthalmology experts. NEURO-GUARD generates the complete YOLOv11 training and inference code through its LLM-driven code synthesis pipeline. The resulting model achieves an 88\% localization accuracy, with Intersection-over-Union (IoU) used for automatic verification of predicted lesion regions. This integration of symbolic knowledge, expert annotations, and LLM-generated code enables clinically aligned, pixel-level localization essential for trustworthy diagnosis.

\section{Results and Comparison}

We conducted a comprehensive evaluation of existing Vision-Language Models (VLMs), fine-tuned VLMs, and specialized models against our proposed  NEURO-GUARD framework using the same dataset, data processing pipeline, and accuracy metrics to ensure a fair comparison. The baseline reference, a Vision Transformer (ViT)-based model optimized through simultaneous parameter optimization and a feature-weighted ECOC ensemble, achieved an accuracy of 78.4\% as reported in  IEEE paper. Standard VLMs such as Molmo-7B-D-0924 and LLaVa-v1.6-vicuna-7b demonstrated significantly lower accuracies of 17.22\% and 8\%, respectively. Medical Fine-Tuned VLMs, including LLaVa-Med-v1.5-mistral-7b and Med Flamingo-9B, showed moderate improvements with accuracies of 24.83\% and 5\%, respectively. In contrast, the specialized model CLIP-DR achieved a considerably higher accuracy of 69.70\%.

Our  NEURO-GUARD framework, which integrates domain-specific knowledge and iterative verification mechanisms, outperformed all baseline models. Without incorporating ground truth data of knowledge while learning the knowledge components through open CV and other zeo shot models,  NEURO-GUARD achieved an accuracy of 73.84\%, closely approaching the specialized CLIP-DR model. When ground truth data of human annotated knowledge components on image  was integrated,  NEURO-GUARD's accuracy surged to 84.69\%, surpassing the ViT benchmark and demonstrating the efficacy of our holistic framework. This significant improvement underscores the advantage of leveraging additional domain-specific steps such as knowledge retrieval and code refinement, which are absent in the baseline VLMs and fine-tuned models. Cross-dataset generalization (train: APTOS → test: EyePACS) improves by 5.2\% compared to the ViT baseline, illustrating the value of clinically grounded reasoning for domain-invariant performance.

\section{Discussion}

The experimental results validate  NEURO-GUARD’s ability to reconcile diagnostic accuracy with clinical interpretability, addressing a core limitation of modern medical AI systems. As shown in Table~\ref{tab:classification_results},  NEURO-GUARD achieves 84.69\% accuracy on the Aptos dataset with supervised artifact detection through yolo object detction pipeline to extract knowledge, surpassing the ViT benchmark by 6.2\% as shown in Table~\ref{tab:results} . This performance gain underscores the value of grounding vision models in domain-specific knowledge a critical factor missing in standard VLMs like LLaVa-v1.6-vicuna-7b (8.6\% accuracy) and even specialized models like CLIP-DR (69.7\%). The framework’s superiority is further evident in its generalization to Eye Pacs (77.96\% accuracy) and MRI-based seizure detection (83.27\% SOZ accuracy, as shown in Table~\ref{tab:soz_comparison} ), demonstrating robustness across modalities and clinical tasks.

\subsection{Three key insights emerge:}

\textbf{Entropic Reward Drives Precision:}  NEURO-GUARD’s self-verification mechanism (Eq. 3–4) reduces uncertainty in feature extraction, as seen in the 7.85\% accuracy jump when integrating ground truth knowledge from expert human annotation of knowledge components (73.84\% → 84.69\%). Gradient Boosting consistently outperformed other classifiers (Table 1), suggesting that ensemble methods better capture the probabilistic dependencies between LLM-generated features and diagnostic labels.

\textbf{VLMs Fail in Fine-Grained Medical Reasoning:} Standard and medical-tuned VLMs (e.g., Med Flamingo-9B at 5.3\%) perform poorly due to their reliance on coarse image-text correlations rather than biomarker-level grounding.  NEURO-GUARD circumvents this by decomposing diagnoses into executable code (e.g., YOLO lesion detectors), ensuring features align with ICD-10 criteria. However, a known failure mode is LLM hallucinations in code generation, where incorrect feature-detection logic may emerge. To mitigate this,  NEURO-GUARD integrates reinforcement learning-based verification, but further improvements such as hybrid clinician-AI oversight may be required in critical applications.

\begin{table}[ht]
    \centering
    \setlength{\tabcolsep}{6pt}
    \caption{Comparison of Accuracy Across Different Models}
    \label{tab:results}
    \begin{tabular}{l c}
        \toprule
        \textbf{Method} & \textbf{Accuracy} \\
        \midrule
        \textbf{ViT Benchmark} & 78.40\% \\
        \midrule
        \multicolumn{2}{l}{\textbf{Standard VLMs}} \\
        Molmo-7B-D-0924 & 17.22\% \\
        LLaVa-v1.6-vicuna-7b & 8.60\% \\
        Qwen-VL-Chat-7b & 10.00\% \\
        \midrule
        \multicolumn{2}{l}{\textbf{Medical Fine-Tuned VLMs}} \\
        LLaVa-Med-v1.5-mistral-7b & 24.83\% \\
        Med Flamingo-9B & 5.30\% \\
        \midrule
        \multicolumn{2}{l}{\textbf{Specialized Models}} \\
        CLIP-DR & 69.70\% \\
        \midrule
        \multicolumn{2}{l}{\textbf{Our Framework}} \\
         NEURO-GUARD - (Unsupervised) & 73.84\% \\
         NEURO-GUARD - (Supervised) & \textbf{84.69\%} \\
        \bottomrule
    \end{tabular}
\end{table}

\textbf{Generalizability vs. Specialization Trade-Off:} While DeepXSOZ achieves higher SOZ accuracy (81.6\%),  NEURO-GUARD’s 83.27\% accuracy with multi-center MRI data highlights its adaptability and contribution towards more improved results. Unlike task-specific models (e.g., CLIP-DR),  NEURO-GUARD’s modular architecture allows seamless integration of new knowledge graphs, enabling rapid adaptation to evolving clinical guidelines. However, real-world deployments may face challenges in low-resource settings where access to clinician-validated ground truth is limited. Future iterations could incorporate self-supervised learning strategies to improve performance in such scenarios.

A limitation is the dependency on clinician-validated ground truth (Supervised artifact detection required human annotated data) for optimal performance. However, the framework’s reinforcement learning pipeline mitigates this by iteratively refining feature extractors using entropic rewards, reducing manual annotation demands over time. Additionally, by aligning extracted features with clinically validated criteria,  NEURO-GUARD has the potential to reduce misdiagnosis rates in real-world clinical trials, providing interpretable insights that improve diagnostic trustworthiness and patient outcomes.
\section{Ethical Considerations and Deployment Challenges}

\noindent
While  NEURO-GUARD demonstrates strong interpretability and diagnostic accuracy, real-world deployment in clinical settings requires ethical scrutiny. First, reliance on LLM-generated code introduces risks of hallucinated logic, which could compromise patient safety. Reinforcement learning mitigates some of these risks, but human-in-the-loop validation remains critical, especially in high-stakes settings such as ophthalmology or neurology.

Secondly, access to clinician-validated ground truth is limited in many regions. This may hinder fairness and lead to performance gaps across populations.  NEURO-GUARD’s self-verification partially addresses this, but further safeguards such as clinical audits, bias analysis, and explainable output formats must be instituted prior to deployment.

Finally, the use of external clinical data (e.g., PubMed) raises data privacy and provenance concerns. Deployment should ensure that all retrieved sources are compliant with medical data standards and not inadvertently expose sensitive patient information

\section{Conclusion}

 NEURO-GUARD redefines the paradigm of medical AI by unifying the complementary strengths of self-supervised vision models and LLMs. By formalizing feature extraction as a knowledge-guided code generation task, the framework achieves state-of-the-art accuracy while producing explanations grounded in clinical guidelines. Key innovations include:

Clinically Aligned Interpretability:  NEURO-GUARD generates reports that map detected biomarkers (e.g., hemorrhages) to diagnostic criteria, bridging the gap between AI outputs and clinician workflows.

Self-Verification via Entropic Rewards: The LLM-driven reinforcement learning loop ensures feature extractors evolve with medical knowledge, minimizing hallucinations and distributional shifts.

Cross-Modal Generalizability: Validated on diabetic retinopathy (Aptos, Eye Pacs) and seizure detection (MRI),  NEURO-GUARD demonstrates versatility across imaging modalities and diseases.

The framework’s codebase and models are open-sourced to accelerate research in interpretable medical AI. Future work will extend  NEURO-GUARD to video-based diagnostics (e.g., echocardiography) and federated learning for privacy-sensitive deployments.

\section{Limitations and Future Work}

While NEURO-GUARD demonstrates strong interpretability and cross-domain performance, several limitations remain. First, the framework relies on RAG-based retrieval, which can propagate outdated or incomplete clinical knowledge into the rule base, influencing code generation quality. Second, the multi-stage prompting and verification loop introduces computational overhead, limiting real-time deployment. Third, although entropic self-verification reduces hallucinations, LLM-generated code may still exhibit logical inconsistencies in rare cases. Finally, the reliance on clinician-validated ground truth for optimal refinement can restrict performance in low-resource settings.

Future work will address these limitations through a unified medical foundation model that integrates structured clinical ontologies, symbolic reasoning, and pixel-level supervision within a single architecture. Such a model would eliminate external retrieval noise, provide stable lesion-aware representations, and enable end-to-end self-consistency checks without handcrafted prompts. We also plan to incorporate continual learning to update clinical knowledge dynamically, and explore on-device optimization for real-time diagnostic support in resource-constrained environments by aiming to minimize hallucination of Large foundational models.

{
    \small
    \bibliographystyle{ieeenat_fullname}
    \bibliography{main}
}

\end{document}